\documentclass[]{article}
\usepackage[letterpaper]{geometry}
\usepackage{mtsummit2015}
\usepackage{times}
\usepackage{url}
\usepackage{latexsym}
\usepackage{natbib}
\usepackage{layout}
\usepackage{graphicx,url,color}
\usepackage{amsmath,amssymb,amsopn}
\usepackage{algorithm,algpseudocode,algorithmicx}
\usepackage{booktabs}
\usepackage{subcaption}
\usepackage[textwidth=0.75in,textsize=script]{todonotes}
\usepackage{gnuplot-lua-tikz}
\usepackage{multirow}

\DeclareMathOperator*{\argmax}{arg\,max}
\DeclareMathOperator*{\argmin}{arg\,min}
\newtheorem{thm}{Theorem}

\providecommand{\norm}[1]{\lVert#1\rVert}
\newcommand{\wtop}{w^{\top}}
\newcommand{\Y}{\mathcal{Y}}
\newcommand{\X}{\mathcal{X}}
\newcommand{\E}{\mathbb{E}}

\begin{document}

% \mtsummitHeader{x}{x}{xxx-xxx}{2015}{45-character paper description goes here}{Author(s) initials and last name go here}

\title{\bf Bandit Structured Prediction for Learning from Partial Feedback in Statistical Machine Translation
}

\author{
\name{\bf Artem Sokolov} \hfill  \addr{sokolov@cl.uni-heidelberg.de}\\ 
        \name{\bf Stefan Riezler$^\ast$} \hfill \addr{riezler@cl.uni-heidelberg.de}\\ 
        \addr{Computational Linguistics and IWR$^\ast$\\ Heidelberg University, 69120 Heidelberg, Germany}
\AND
       \name{\bf Tanguy Urvoy} \hfill \addr{tanguy.urvoy@orange.com}\\
       \addr{Orange Labs, 2 Avenue Pierre Marzin, 22307 Lannion, France}
}

\maketitle
\pagestyle{empty}

\begin{abstract} We present an approach to structured prediction from bandit feedback, called \emph{Bandit Structured Prediction}, where only the value of a task loss function at a single predicted point, instead of a correct structure, is observed in learning. We present an application to discriminative reranking in Statistical Machine Translation (SMT) where the learning algorithm only has access to a $1- \textrm{BLEU}$ loss evaluation of a predicted translation instead of obtaining a gold standard reference translation. In our experiment bandit feedback is obtained by evaluating BLEU on reference translations without revealing them to the algorithm. This can be thought of as a simulation of interactive machine translation where an SMT system is personalized by a user who provides single point feedback to predicted translations. Our experiments show that our approach improves translation quality and is comparable to approaches that employ more informative feedback in learning.
\end{abstract}

\section{Introduction}
Learning from \emph{bandit}\footnote{The name is inherited from a model where in each round a gambler pulls an arm of a different slot machine (``one-armed bandit''), with the goal of maximizing his reward relative to the maximal possible reward, without apriori knowledge of the optimal slot machine.} feedback describes an online learning scenario, where on each of a sequence of rounds, a learning algorithm makes a prediction, and receives partial information in terms of feedback to a single predicted point. In difference to the full information supervised scenario, the learner does not know what the correct prediction looks like, nor what would have happened if it had predicted differently. This scenario has (financially) important real world applications such as online advertising \citep{ChapelleETAL:14} that showcases a tradeoff between exploration (a new ad needs to be displayed in order to learn its click-through rate) and exploitation (displaying the ad with the current best estimate is better in the short term). Crucially, in this scenario it is unrealistic to expect more detailed feedback than a user click on the displayed ad. 
Similar to the online advertising scenario, there are many potential applications of bandit learning to NLP situations where feedback is limited for various reasons. For example, online learning has been applied successfully in interactive statistical machine translation (SMT) \citep{BertoldiETAL:14,DenkowskiETAL:14,GreenETAL:14}. Post-editing feedback clearly is limited by its high cost and by the required expertise of users, however, current approaches force the full information supervised scenario onto the problem of learning from user post-edits. Bandit learning would allow to learn from partial user feedback that is easier and faster to obtain than full information.
An example where user feedback is limited by a time constraint is simultaneous translation of a speech input stream \citep{ChoETAL:13}. Clearly, it is unrealistic to expect user feedback that goes beyond a one-shot user quality estimate of the predicted translation in this scenario.
Another example is SMT domain adaptation where the translations of a large out-of-domain model are re-ranked based on bandit feedback on in-domain data. This can also be seen as a simulation of personalized machine translation where a given large SMT system is adapted to a user solely by single-point user feedback to predicted structures. 

The goal of this paper is to develop algorithms for structured prediction from bandit feedback, tailored to NLP problems. We investigate possibilities to ``banditize'' objectives such as expected loss \citep{Och:03,SmithEisner:06,GimpelSmith:10} %or cross-entropy loss \cite{GreenETAL:14}
that have been proposed for structured prediction in NLP. Since most current approaches to bandit optimization rely on a multiclass classification scenario, the first challenge of our work is to adapt bandit learning to structured prediction over exponentially large structured output spaces \citep{TaskarETAL:04,TsochantaridisETAL:05}. Furthermore, most theoretical work on online learning with bandit feedback relies on convexity assumptions about objective functions, both in the non-stochastic adversarial setting \citep{FlaxmanETAL:05,Shalev-Shwartz:12} as well as in the stochastic optimization framework \citep{Spall:03,NemirovskiETAL:09,BachMoulines:11}. Our case is a non-convex optimization problem, which we analyze in the simple and elegant framework of pseudogradient adaptation that allows us to show convergence of the presented algorithm \citep{PolyakTsypkin:73,Polyak:87}. %and give convergence rates for suitable (strong) convexifications of objective functions \cite{PolyakTsypkin:73,Polyak:76,Polyak:87}.

The central contributions of this paper are:
\begin{itemize}
\item An algorithm for minimization of expected loss %and cross-entropy loss
for structured prediction from bandit feedback, called \emph{Bandit Structured Prediction}. 
\item An analysis of convergence %and convergence rates
of our algorithm in the stochastic optimization framework of pseudogradient adaptation.
\item An experimental evaluation %on standard multiclass and structured datasets
on structured learning in SMT. Our experiment follows a simulation design that is standard in bandit learning, namely by simulating bandit feedback by evaluating task loss functions against gold standard structures without revealing them to the learning algorithm. %We show the efficacy of our algorithm on different losses of increasing complexity (0/1 loss, Hamming loss, BLEU), and in comparison to existing generic bandit algorithms for multiclass classification \cite{KakadeETAL:08} and ranking \cite{YueJoachims:09}.
\end{itemize}

As a disclaimer, we would like to note that improvements over traditional full-information structured prediction cannot be expected from learning from partial feedback. Instead, the goal is to investigate learning situations in which full information is not available. Similarly, a comparison between our approach and dueling bandits \citep{YueJoachims:09} is skewed towards the latter approach that has access to two-point feedback instead of one-point feedback as in our case. While it has been shown that querying the loss function at two points leads to convergence results that closely resemble bounds for the full information case \citep{AgarwalETAL:10}, such feedback is clearly twice as expensive and, depending on the application, might not be elicitable from users.

\section{Related Work}
\paragraph{Stochastic Approximation.} Online learning from bandit feedback dates back to \cite{Robbins:52} who formulated the task as a problem of sequential decision making. His analysis, as ours, is in a stochastic setting, i.e., certain assumptions are made on the probability distribution of feedback and its noisy realization. Stochastic approximation covers bandit feedback as noisy observations which only allow to compute noisy gradients that equal true gradients in expectation. While the stochastic approximation framework is quite general, most theoretical analyses of convergence and convergence rate are based on (strong) convexity assumptions \citep{PolyakJuditsky:92,Spall:03,NemirovskiETAL:09,BachMoulines:11,BachMoulines:13} and thus not applicable to our case.

\paragraph{Non-Stochastic Bandits.} \cite{AuerETAL:02} initiated an active area of research on non-stochastic bandit learning, i.e., no statistical assumptions are made on the distribution of feedback, including models of feedback as a malicious choice of an adaptive adversary. The adversarial bandit setting has been extended to take  { context} or { side information} into account, using models based on general linear classifiers \citep{AuerETAL:02,LangfordEtAl:07,ChuEtAl:11}. However, they formalize a multi-class classification problem that is not easily scalable to general exponentially large structured output spaces. Furthermore, most theoretical analyses rely on online (strongly) convex optimization \citep{FlaxmanETAL:05,Shalev-Shwartz:12} thus limiting the applicability to our case.

\paragraph{Neurodynamic Programming.} \cite{BertsekasTsitsiklis:96} cover optimization for neural networks and reinforcement learning under the name of ``neurodynamic programming''. Both areas are dealing with non-convex objectives that lead to stochastic iterative algorithms. Interestingly, the available analyses of non-convex optimization for neural networks and reinforcement learning in \cite{BertsekasTsitsiklis:96}, \cite{SuttonETAL:00}, or \cite{Bottou:04} rely heavily on \cite{PolyakTsypkin:73}'s pseudogradient framework. We apply their simple and elegant framework directly to give asymptotic guarantees for our algorithm. %show convergence and convergence rates of our algorithms.

\paragraph{NLP Applications.} In the area of NLP, recently algorithms for response-based learning have been proposed to alleviate the supervision problem by extracting supervision signals from task-based feedback to system predictions. For example, \cite{GoldwasserRoth:13} presented an online structured learning algorithm that uses positive executability of a semantic parse against a database to convert a predicted parse into a gold standard structure for learning. \cite{RiezlerETAL:14} apply a similar idea to SMT by using the executability of a semantic parse of a translated database query as signal to convert a predicted translation into gold standard reference in structured learning. \cite{SokolovETAL:15} present a coactive learning approach to structured learning in SMT where instead of a gold standard reference a slight improvement over the prediction is shown to be sufficient for learning. \cite{SalujaZhang:14} present an incorporation of binary feedback into an latent structured SVM for discriminative SMT training. NLP applications based on reinforcement learning have been presented by \cite{BranavanETAL:09} or \cite{ChangETAL:15}. Their model differs from ours in that it is structured as a sequence of states at which actions and rewards are computed, however, the theoretical foundation of both types of models can be traced back to \cite{PolyakTsypkin:73}'s pseudogradient framework .

\section{Expected Loss Minimization under Full Information}
\label{sec:exprisk}

The expected loss learning criterion for structured prediction is defined as a minimization of
the expectation of a task loss function with respect to the conditional distribution
over structured outputs \citep{GimpelSmith:10,YuilleHe:12}. More formally, let
$\X$ be a structured input space, let $\Y(x)$ be the set of possible output
structures for input $x$, and let $\Delta_y: \Y  \rightarrow [0,1]$ quantify the
loss $\Delta_y(y')$ suffered for making errors in predicting $y'$ instead of
$y$; as a rule, $\Delta_y(y')=0$ iff $y=y'$. 
Then, for a data distribution $p(x,y)$, the learning criterion is defined as minimization of the expected  loss
\begin{align}
\label{eq:exp-risk} 
\E&_{ p(x,y) p_w(y'|x)}  \left[ \Delta_y(y') \right]
= \sum_{x,y}  p(x,y) \sum_{y' \in \Y(x)} \Delta_y(y') p_w(y'|x).
\end{align}
Assume further that output structures given inputs are distributed according to an underlying Gibbs distribution (a.k.a. conditional exponential or log-linear model)
\begin{equation}
\label{eq:llm}
p_w(y|x) = \exp(\wtop \phi(x,y))/Z_w(x), \notag
\end{equation}
where $\phi: \X\times\Y\rightarrow\mathbb{R}^d$ is a joint feature representation of inputs and outputs, $w \in \mathbb{R}^d$ is a weight vector, and $Z_w(w)$ is a normalization constant.

The natural rule for prediction or inference is according to the minimum Bayes risk principle
\begin{equation}
\label{eq:mbr}
\hat y_w(x) = \argmin_{y \in \Y(x)} \sum_{y'\in \Y(x)} \Delta_y(y') p_w(y'|x).
\end{equation}
This requires an evaluation of $\Delta_y(y')$ over the full output space, which is standardly avoided in practice by performing inference according to a maximum a posteriori (MAP) criterion (which equals criterion~\eqref{eq:mbr} for the special case of $\Delta_y(y')=\mathbf{1}[y \neq y']$ where $\mathbf{1}[s]$ evaluates to $1$ if statement $s$ is true, $0$ otherwise)

\begin{align}
\label{eq:linear} 
\hat{y}_w(x) &= \argmax_{y \in \Y(x)} p_w(y|x)\\  
& = \argmax_{y \in \Y(x)} \wtop \phi(x,y). \notag
\end{align}
 
Furthermore, since it is unfeasible to take expectations over the full space $\X \times \Y$ to perform minimization of objective \eqref{eq:exp-risk}, in the full information case the data distribution $p(x,y)$ is approximated by the empirical distribution $\tilde p(x,y) = \frac{1}{T} \sum_{t=0}^T \mathbf{1}[x = x_t]\mathbf{1}[y = y_t ]$ for i.i.d. training data  $\{(x_t,y_t)\}_{t=0}^T$. This yields the objective
\begin{align}
\label{eq:emp-exp-risk}
\E&_{ \tilde{p}(x,y)  p_w(y'|x)}  \left[ \Delta_y(y') \right] 
= \frac{1}{T}\sum_{t=0}^T \sum_{y' \in \Y(x_t)} \Delta_{y_t}(y') p_w(y'|x_t).
\end{align}

While being continuous and differentiable, the expected loss criterion is typically non-convex. For example, in SMT, expected loss training for the standard task loss BLEU leads to highly non-convex optimization problems. Despite of this, most approaches rely on gradient-descent techniques for optimization (see \cite{Och:03}, \cite{SmithEisner:06}, \cite{HeDeng:12},  \cite{AuliETAL:14}, \cite{WuebkerETAL:15}, \emph{inter alia}) by following the opposite direction of the gradient of \eqref{eq:emp-exp-risk}:
\begin{align*}
%\label{eq:grad-exp-risk}
\nabla & \E_{\tilde{p}(x,y) p_w(y'|x)} \left[ \Delta_y(y') \right] \\ \notag
&= \E_{\tilde{p}(x,y)} \Big[ \E_{p_w(y'|x)} [ \Delta_y(y') \phi(x,y') ] 
-\E_{p_w(y'|x)} [ \Delta_y(y') ] \; \E_{p_w(y'|x)} [  \phi(x,y') ]\Big] \\ \notag
 & =\E_{\tilde{p}(x,y) p_w(y'|x)} \Big[ \Delta_y(y') ( \phi(x,y') - \E_{p_w(y'|x)} [ \phi(x,y') ] ) \Big]. \notag
\end{align*}

\section{Bandit Structured Prediction}
Bandit feedback in structured prediction means that the gold standard output structure $y$, with respect to which the objective function is evaluated, is not revealed to the learner. Thus we can neither calculate the gradient of the objective function \eqref{eq:emp-exp-risk} nor evaluate the task loss $\Delta$ as in the full information case. A solution to this problem is to pass the evaluation of the loss function to the user, i.e, we access the loss directly through user feedback without assuming existence of a fixed reference $y$. We indicate this by dropping the subscript $y$ in $\Delta(y')$. Assuming a fixed, unknown distribution $p(x)$ over input structures, we can formalize the following new objective for expected loss minimization in a bandit setup
\begin{align}
\label{eq:bandit-exp-risk}
%TU
J(w)&=
\E_{ p(x) p_w(y'|x)}  \left[ \Delta(y') \right]\\
&= \sum_{x}  p(x) \sum_{y' \in \Y(x)} \Delta(y') p_w(y'|x). \notag
\end{align}
Optimization of this objective is then as follows:
\begin{enumerate}
\item We assume a sequence of input structures $x_t, t= 1, \ldots, T$ that are generated by a fixed, unknown distribution $p(x)$.
\item We use a Gibbs distribution estimate as a sampling distribution to perform simultaneous exploration / exploitation on output structures \citep{AbernethyRakhlin:09}.
\item We use feedback to the sampled output structures to construct a parameter update rule that is an unbiased estimate of the true gradient of objective \eqref{eq:bandit-exp-risk}.
\end{enumerate}

\subsection{Algorithm}
\begin{algorithm}[t]
\caption{Bandit Structured Prediction}
\label{alg:bandit}
\begin{algorithmic}[1]
\algnotext{EndFor}
\State Input: sequence of learning rates $\gamma_t$ \label{alg:line:lrates}
\State Initialize $w_0$ \label{alg:line:w0}
\For{$t=0,\ldots,T$}
\State Observe $x_t$
\State Calculate $\E_{p_{w_t}(y'|x_t)}[\phi(x_t,y')]$ \label{alg:line:avg}
\State Sample $\tilde y_t \sim p_{w_t}(y'|x_t)$ \label{alg:line:sample}
\State Obtain feedback $\Delta(\tilde y_t)$ 
\State Update $w_{t+1} = w_t - \gamma_t \; \Delta(\tilde y_t)(\phi(x_t,\tilde y_t) -\E_{p_{w_t}(y'|x_t)}[\phi(x_t,y')])$  \label{alg:line:update}
\EndFor
\end{algorithmic}
\end{algorithm}

Algorithm \ref{alg:bandit} implements these ideas as follows: We assume as
input a given learning rate schedule (line~\ref{alg:line:lrates}) and a
deterministic initialization $w_0$ of the weight vector
(line~\ref{alg:line:w0}). 
For each random i.i.d. input structure $x_t$, we calculate the expected feature
count (line \ref{alg:line:avg}). This can be done exactly, provided the
underlying graphical model permits a tractable calculation, or for intractable models, with MCMC sampling.  We then sample an output structure $\tilde y_t$ from the Gibbs model (line \ref{alg:line:sample}).
If the number of output options is small, this is done by  sampling from a multinomial distribution. Otherwise, we use a Perturb-and-MAP approach \citep{PapandreouYuille:11}, restricted to unary potentials, to obtain an approximate Gibbs sample
without waiting for the MC chain to mix.  Finally, an update in the negative
direction of the instantaneous gradient, evaluated at the input structure $x_t$
(line \ref{alg:line:update}), is  performed. 

Intuitively, the algorithm compares the sampled feature vector to the average
feature vector, and performs a step into the opposite direction of this
difference, the more so the higher the loss of the sampled structure is. In the extreme case, if the
sampled structure is correct ($\Delta(\tilde y_t) = 0$), no update is performed.

\subsection{Stochastic Approximation Analysis}

The construction of the update in Algorithm \ref{alg:bandit} as a stochastic realization of the true gradient allows us to analyze the algorithm as a stochastic approximation algorithm.  We show how our case can be fit in the pseudogradient adaptation framework of \cite{PolyakTsypkin:73} which gives asymptotic guarantees for non-convex and convex objectives. They characterize an iterative process 
\begin{align}
\label{eq:process}
w_{t+1} = w_t - \gamma_t \; s_t
\end{align}
where $\gamma_t \geq 0$ is a learning rate, $w_t$ and $s_t$ are vectors in $\mathbb{R}^d$ with fixed $w_0$, and the distribution of $s_t$ depends on $w_0, \ldots, w_t$. For a given lower bounded and differentiable function $J(w)$ with Lipschitz continuous gradient $\nabla J(w)$, that is, for all $w, w'$, there exists $L \geq 0$, such that
\begin{align}
\label{eq:lipschitz}
\norm{\nabla J(w+w') - \nabla J(w)} \leq L \norm{w'},
\end{align}
%and minimum $\inf J(w) = 0$, 
the vector $s_t$ in process \eqref{eq:process} is said to be a \emph{pseudogradient} of $J(w)$ if
\begin{align}
\label{eq:pseudogradient}
\nabla J(w_t)^{\top} \E[s_t] \geq 0,
\end{align}
where the expectation is taken over all sources of randomness. 
Intuitively, the pseudogradient $s_t$ is on average at an acute angle with the true gradient, meaning that $-s_t$ is on average a direction of decrease of the functional $J(w)$. 

In order to show convergence of the iterative process \eqref{eq:process}, besides conditions \eqref{eq:lipschitz} and \eqref{eq:pseudogradient}, only mild conditions on boundedness of the pseudogradient 
\begin{align}
\label{eq:growth}
\E[ \norm{s_t}^2] < \infty,
\end{align}
and on the use of a decreasing learning rate satisfying
\begin{align}
\label{eq:learningrate}
\gamma_t \geq 0, \; \sum_{t=0}^{\infty} \gamma_t = \infty, \;  \sum_{t=0}^{\infty} \gamma_t^2 < \infty,
\end{align}
are necessary. Under the exclusion of trivial solutions such as $s_t =  \boldsymbol{0}$, the following convergence assertion can be made:
\begin{thm}[\cite{PolyakTsypkin:73}, Thm. 1]
\label{thm:polyak}
Under conditions \eqref{eq:lipschitz}--\eqref{eq:learningrate}, for any $w_0$ in process \eqref{eq:process}:
\begin{align*}
J(w_t) \rightarrow J^\ast \; \textrm{ a.s., and } \lim_{t \rightarrow \infty} \nabla J(w_t)^{\top} \E(s_t) = 0.
\end{align*}
\end{thm}
The significance of the theorem is that its conditions can be checked easily, and it applies to a wide range of cases, including non-convex functions, in which case the convergence point $J^\ast$ is a critical point of $J(w)$.

The convergence analysis of Theorem \ref{thm:polyak} can be applied to Algorithm \ref{alg:bandit} as follows: First note that we can define our functional $J(w)$ with respect to expectations over the full space of $\X$ as $J(w) = \E_{p(x)p_w(y'|x)}[\Delta(y')]$. 
This means, convergence of the algorithm can be understood directly as a generalization result that extends to unseen data. In order to show this result, we have to verify conditions \eqref{eq:lipschitz}--\eqref{eq:learningrate}. It is easy to show that condition \eqref{eq:lipschitz} holds for our functional $J(w)$. Next we match the update in Algorithm 1 to a vector
\begin{align*}
s_t = \Delta(\tilde y_t)(\phi(x_t,\tilde y_t) -\E_{p_{w_t}(y'|x_t)}[\phi(x_t,y')]).
\end{align*}
Taking the expectation of $s_t$ yields $\E_{p(x)p_{w_t}(y'|x)}[s_t] = \nabla J(w_t)$ such that condition \eqref{eq:pseudogradient} is satisfied by 
\begin{align*}
\nabla J(w_t)^{\top} \E_{p(x)p_{w_t}(y'|x)}[s_t] = \norm{\nabla J(w_t)}^2 \geq 0.
\end{align*}
Assuming $\norm{\phi(x,y')} \leq R$ and $\Delta(y') \in [0,1]$ for all $x,y'$, condition \eqref{eq:growth} is satisfied by
\begin{align*}
\E_{p(x)p_{w_t}(y'|x)}[ \norm{s_t}^2]\leq 4R^2.
\end{align*}
For a decreasing learning rate, e.g.,  $\gamma_t = 1/t$, condition \eqref{eq:learningrate} holds, such that convergence to a critical point of the expected risk follows according to Theorem \ref{thm:polyak}.

\section{Structured Dueling Bandits}
%\subsection{Dueling Bandits for Pairwise Ranking}

For purposes of comparison, we present an extension of \cite{YueJoachims:09}'s dueling bandits algorithm to structured prediction problems. The original algorithm is not specifically designed for structured prediction problems, but it is generic enough to be applicable to such problems %, respectively, when the number of output structures is small or 
when the quality of a parameter vector can be proxied through loss evaluation of an inferred structure.

\begin{algorithm}[t]
\renewcommand{\thealgorithm}{}
\caption{Structured Dueling Bandits}
\label{alg:dueling}

\begin{algorithmic}[1]
\algnotext{EndFor}
\algnotext{EndIf}

\State Input: $\gamma, \delta, w_0$
\For{$t=0,\ldots,T$}
\State Observe $x_t$
\State Sample unit vector $u_t$ uniformly
\State Set $w'_t = w_t + \delta u_t$\label{alg:line:dueldelta}
\State Compare $\Delta(\hat y_{w_t}(x_t))$ to $\Delta(\hat y_{w'_t}(x_t))$ \label{alg:line:compare}%the quality of $w_t$ and $w'_t$
\If{$w'_t$ wins}
\State $w_{t+1} = w_t + \gamma u_t$\label{alg:line:duelgamma}
\Else
\State $w_{t+1} = w_t$
\EndIf
\EndFor

\end{algorithmic}
\end{algorithm}

The Structured Dueling Bandits algorithm compares a current weight vector $w_t$ with a
neighboring point $w'_t$ along a direction $u_t$, performing exploration
(controlled by $\delta$, line~\ref{alg:line:dueldelta}) by probing random directions, and exploitation (controlled by $\gamma$,
line~\ref{alg:line:duelgamma}) by taking a step into the winning direction. The comparison step in
line~\ref{alg:line:compare} is adapted to structured prediction from the original algorithm of \cite{YueJoachims:09} by comparing the quality of $w_t$ and $w'_t$ via an evaluation of the losses
$\Delta(\hat y_{w_t}(x_t))$ and $\Delta(\hat y_{w'_t}(x_t))$ of the structured arms corresponding to MAP prediction \eqref{eq:linear} under $w_t$ and $w'_t$, respectively.

Further, note that the Structured Dueling Bandit algorithm requires access to a two-point feedback instead of a one-point feedback as in case of Bandit Structured Prediction (Algorithm~1). It has been shown that two-point feedback leads to convergence results that are close to those for learning from full information \cite{AgarwalETAL:10}. However, two-point feedback is twice as expensive as one-point feedback, and most importantly, such feedback might not be elicitable from users in real-world situations where feedback is limited by time- and resource-constraints. This limits the range of applications of Dueling Bandits to real-world interactive scenarios.

\section{Experiments}
Our experimental design follows the standard of simulating bandit feedback by evaluating task loss functions against gold standard structures without revealing them to the learner. We compare the proposed Structured Bandit Prediction algorithm to Structured Dueling Bandits, and report results by test set evaluations of the respective loss functions under MAP inference. Furthermore, we evaluate models at different iterations according to their loss on the test set in order to visualize the empirical convergence behavior of the algorithms.

All experiments with bandit algorithms perform online learning for parameter estimation, and apply early stopping to choose the last model in a learning sequence for online-to-batch conversion at test time. Final results for bandit algorithms are averaged over 5 independent runs.

%\subsection{SMT Reranking under 1-BLEU Loss}

In this experiment, we present bandit learning for the structured $1 - \textrm{BLEU}$ loss used in SMT. The setup is a reranking approach to SMT domain adaptation where the $k$-best list of an out-of-domain model is re-ranked (without re-decoding) based on bandit feedback from in-domain data. This can also be seen as a simulation of personalized machine translation where a given large SMT system is adapted to a user solely by single-point user feedback to predicted structures. 

We use the data from the WMT 2007 shared task for domain adaptation experiments in a popular benchmark setup from Europarl to NewsCommentary for French-to-English \citep{KoehnSchroeder:07,DaumeETAL:11}. We tokenized and lowercased our data using the \texttt{moses} toolkit, and prepared word alignments by \texttt{fast\_align} \citep{DyerETAL:13}. The SMT setup is phrase-based translation using non-unique 5,000-best lists from \texttt{moses} \citep{KoehnETAL:07} and a 4-gram language model \citep{HeafieldETAL:13}. 

The \emph{out-of-domain} baseline SMT model is trained on 1.6 million parallel Europarl data and includes the English side of Europarl and \emph{in-domain} NewsCommentary in the language model. The model uses 15 dense features (6 lexicalized reordering features, 1 distortion, 1 out-of-domain and 1 in-domain language model, 1 word penalty, 5 translation model features) that are tuned with MERT \citep{Och:03} on a dev set of Europarl data (\texttt{dev2006}, 2,000 sentences). 
The full-information \emph{in-domain} SMT model gives an upper bound by MERT tuning the out-of-domain model on in-domain development data (\texttt{nc-dev2007}, 1,057 sentences).
MERT runs for both baseline models were repeated 7 times and median results are reported.

%\begin{table}[t]
%\centering
%%\resizebox{\columnwidth}{!}{
%\begin{tabular}{cc|cc}
%\toprule
% \multicolumn{2}{c|}{full information} & \multicolumn{2}{c}{bandit information}\\
% \bf in-domain SMT & \bf out-domain SMT & \bf DuelingBandit & \bf BanditStruct\\
%\midrule
% 0.2868 & 0.2509 & 0.2693$_{\pm 0.0001}$ & 0.265$_{\pm 0.001}$ \\
%\bottomrule
%\end{tabular}
%%}
%\caption{Corpus BLEU (under MAP decoding) on test set for SMT domain adaptation from Europarl to NewsCommentary by $k$-best reranking.}
%\label{tab:smt}
%\end{table}

\begin{table}[t]
\centering
%\resizebox{\columnwidth}{!}{
\begin{tabular}{cc|cc}
\toprule
 \multicolumn{2}{c|}{full information} & \multicolumn{2}{c}{bandit information}\\
 \bf in-domain SMT & \bf out-domain SMT & \bf DuelingBandit & \bf BanditStruct\\
\midrule
 0.2854 & 0.2579 & 0.2731$_{\pm 0.001}$ & 0.2705$_{\pm 0.001}$ \\
\bottomrule
\end{tabular}
%}
\caption{Corpus BLEU (under MAP decoding) on test set for SMT domain adaptation from Europarl to NewsCommentary by $k$-best reranking.}
\label{tab:smt}
\end{table}

%\begin{figure}[t]
%  \centering
%    \vspace{-2em}
%    \resizebox{\columnwidth}{!}{\noendlineinput{smt_test_noEL.texgp}}
%  \caption{Corpus-BLEU on test set for early stopping at different iterations for the SMT task.\label{fig:smt_test}}
%\end{figure}

\begin{figure}[t]
  \centering
    \vspace{-2em}
    \resizebox{\columnwidth}{!}{\begingroup
    \endlinechar=-1 \input{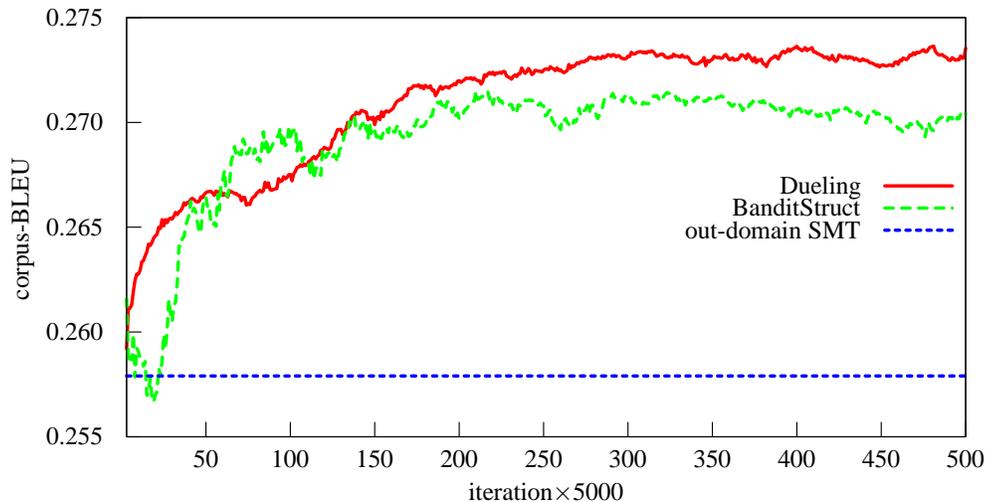}\endgroup}
  \caption{Corpus-BLEU on test set for early stopping at different iterations for the SMT task.\label{fig:smt_test}}
\end{figure}

Learning under bandit feedback started at the learned weights of the \emph{out-of-domain} median model. It uses the parallel NewsCommentary data (\texttt{news-commentary}, 43,194 sentences) to simulate bandit feedback, by evaluating the sampled translation against the gold standard reference using as loss function $\Delta$ a smoothed per-sentence $1 - \textrm{BLEU}$ (by flooring zero $n$-gram counts to $0.01$). The meta-parameters of Dueling Bandits and Bandit Structured Prediction were adjusted by online optimization of cumulative per-sentence $1 - \textrm{BLEU}$ on a small \emph{in-domain} dev set (\texttt{nc-devtest2007}, 1,064 parallel sentences). The final results are obtained by online-to-batch conversion where the model trained for 100 epochs on 43,194 \emph{in-domain} training data is evaluated on a separate \emph{in-domain} test set (\texttt{nc-test2007}, 2,007 sentences).

Table \ref{tab:smt} shows that the results for Bandit Structured Prediction and Dueling Bandits are very close, however, both are significant improvements over the out-of-domain SMT model that even includes an in-domain language model. We show the standard evaluation of the corpus-BLEU metric evaluated under MAP inference. The range of possible improvements is given by the difference of the BLEU score of the in-domain model and the BLEU score of the out-of-domain model -- nearly 3 BLEU points. Bandit learning can improve the out-of-domain baseline by about 1.26 BLEU points (Bandit Structured Prediction) and by about 1.52 BLEU points (Dueling Bandits). 
All result differences are statistically significant at a $p$-value of $0.0001$, using an Approximate Randomization test \citep{RiezlerMaxwell:05,ClarkETAL:11}. Figure \ref{fig:smt_test} shows that per-sentence BLEU is a difficult metric to provide single-point feedback, yielding a non-smooth progression of loss values against iterations for Bandit Structured Prediction. The progression of loss values is smoother and empirical convergence speed is faster for Dueling Bandits since it can exploit preference judgements instead of having to trust real-valued feedback.

\section{Discussion}
We presented an approach to \emph{Bandit Structured Prediction} that is able to learn from feedback in form of an evaluation of a task loss function for \emph{single} predicted structures. Our experimental evaluation showed promising results, both compared to Structured Dueling Bandits that employ two-point feedback, and compared to full information scenarios where the correct structure is revealed.

Our approach shows its strength where correct structures are unavailable and two-point feedback is infeasible. In future work we would like to apply bandit learning to scenarios with limited human feedback such as the interactive SMT applications discussed above.  In such scenarios, per-sentence BLEU might not be the best metric to quantify feedback. We will instead investigate feedback based on HTER \citep{SnoverETAL:06}, or based on judgements according to Likert scales \citep{Likert:32}.

\section*{Acknowledgements}

This research was supported in part by DFG grant RI-2221/2-1 ``Grounding Statistical Machine Translation in Perception and Action''.

\bibliographystyle{apalike}
\bibliography{references}

\end{document}